\documentclass[twocolumn]{article} 

\usepackage[utf8]{inputenc} 
\usepackage[T1]{fontenc} 
\usepackage[english]{babel} 
\usepackage{ifpdf,amsmath,amsthm,amssymb,amsfonts,newtxtext,newtxmath} 
\usepackage{array,graphicx,dcolumn,multirow,abstract,hanging} 
\usepackage[font={it,footnotesize},labelfont=bf]{caption} 
\usepackage[hyperfootnotes=false,breaklinks=true,hidelinks]{hyperref} 
\usepackage[noabbrev,capitalize]{cleveref} 
\usepackage{float} 
\usepackage{url} 
\usepackage[version=4]{mhchem} 
\usepackage{siunitx} 
\usepackage[super,sort&compress,comma]{natbib}  
\usepackage{booktabs} 
\usepackage[normalem]{ulem}
\useunder{\uline}{\ul}{}
\newcolumntype{L}[1]{>{\raggedright\arraybackslash }p{#1}} 
\usepackage{authblk}
\newcolumntype{C}[1]{>{\centering\arraybackslash }p{#1}}
\newcolumntype{R}[1]{>{\raggedleft\arraybackslash }p{#1}}
\newcolumntype{d}[1]{D{.}{.}{#1}} 

\let\tempone\itemize
\let\temptwo\enditemize
\let\tempthree\enumerate
\let\tempfour\endenumerate
\renewenvironment{itemize}{\tempone\setlength{\itemsep}{0pt}}{\temptwo}
\renewenvironment{enumerate}{\tempthree\setlength{\itemsep}{0pt}}{\tempfour}

\date{} 
\title{\textit{Mos-Gen: A Generative Molecular Framework for Mosquito Insecticide Design} } 
\author[1]{Lina Wang}
\author[2]{Yaning Cui}
\author[*3]{Zhifeng Gao}
\author[*1]{Ping Xing}
\author[*1]{Biao Jiang}

\affil[1]{Shanghai Institute of Organic Chemistry, Chinese Academy of Sciences, 345 Lingling Road, Xuhui District, Shanghai, China}
\affil[2]{DP Technology, Shanghai, China}
\affil[3]{DP Technology, Beijing, China, gaozf@dp.tech}

\begin{document}
\twocolumn[ 
\vspace{-.5in}
\maketitle
\vspace{-0.3in}
\begin{center}
\section*{Abstract}
\end{center}
{\large 

Mosquito-borne infectious diseases cause more than 700,000 deaths worldwide each year. The long-term use of conventional chemical insecticides has induced serious resistance problems, creating an urgent need to develop novel, highly effective, and ecologically sustainable alternatives. While existing artificial intelligence approaches in this domain have focused primarily on activity prediction and classification, they leave a critical gap in the de novo generation of novel molecular scaffolds. In this study, we propose Mos-Gen, a motif-aware generate collaborative framework that couples the pretrained molecular representation model Uni-Mol with a variational autoencoder (VAE), specifically tailored for the design of disulfide-containing allicin derivatives as mosquito insecticides. Among the generated candidates, fourteen compounds — comprising nine predicted positives and five predicted negatives — were selected for chemical synthesis and experimental validation. The hit rate among the predicted positives reached 78\%, whereas none of the predicted negatives exhibited mosquitocidal activity. These experimental results fully validated the high-precision screening capability of the Mos-Gen framework.

}
\vspace{0.4in}
]
\setlength{\baselineskip}{11pt plus.2pt}

\section{Introduction} 

Mosquitoes pose a severe global public health threat, serving as vectors for several devastating infectious diseases. Mosquito-borne diseases afflict millions of people annually, causing over 700,000 deaths worldwide\cite{1_Vector-borne}. Malaria, a parasitic infection transmitted by Anopheles mosquitoes, accounts for an estimated 249 million cases globally and results in more than 608,000 deaths each year, with the majority of fatalities occurring in children under five years of age\cite{2,3_satapathy2024japanese}. Dengue, Zika, chikungunya, yellow fever, and Japanese encephalitis represent additional mosquito-borne diseases that contribute substantially to the global burden of illness through high-morbidity outbreaks\cite{4,5}. Mosquito management programmes have relied heavily on the routine application of chemical insecticides, a dependence that has driven the progressive development of resistance among vector populations and raised widespread concerns regarding environmental pollution, biomagnification, and adverse effects on public and animal health\cite{6,7_hillary2024plantproducts,8,9}. The escalating global risk posed by vector-borne diseases, together with their associated morbidity and mortality, underscores the urgent need to develop novel, ecologically sustainable mosquitocidal compounds.\cite{10}

\vspace{0.5em}

\noindent Artificial intelligence has revolutionized drug discovery, compressing traditional decades-long development pipelines into a matter of months.\cite{11,12,13}. A landmark example is the AI-driven development of the first-in-class TNIK inhibitor INS018\_055 (rentosertib) for the treatment of idiopathic pulmonary fibrosis, which progressed from target identification to preclinical candidate nomination in approximately 18 months, fully demonstrating the transformative potential of generative AI-driven drug discovery pipelines\cite{14}. Yet, compared to the drug discovery domain, the application of AI methods in the publicly available pesticide research literature has lagged considerably\cite{15,16,17}. Current AI applications to mosquito insecticides have focused predominantly on discriminative tasks such as larvicidal activity prediction and repellent screening and classification, and have failed to systematically address the de novo generation of novel chemical scaffolds with desired insecticidal profiles.\cite{18,19,20}

\clearpage

\begin{figure}[h!]
\centering
\includegraphics[width=1\columnwidth]{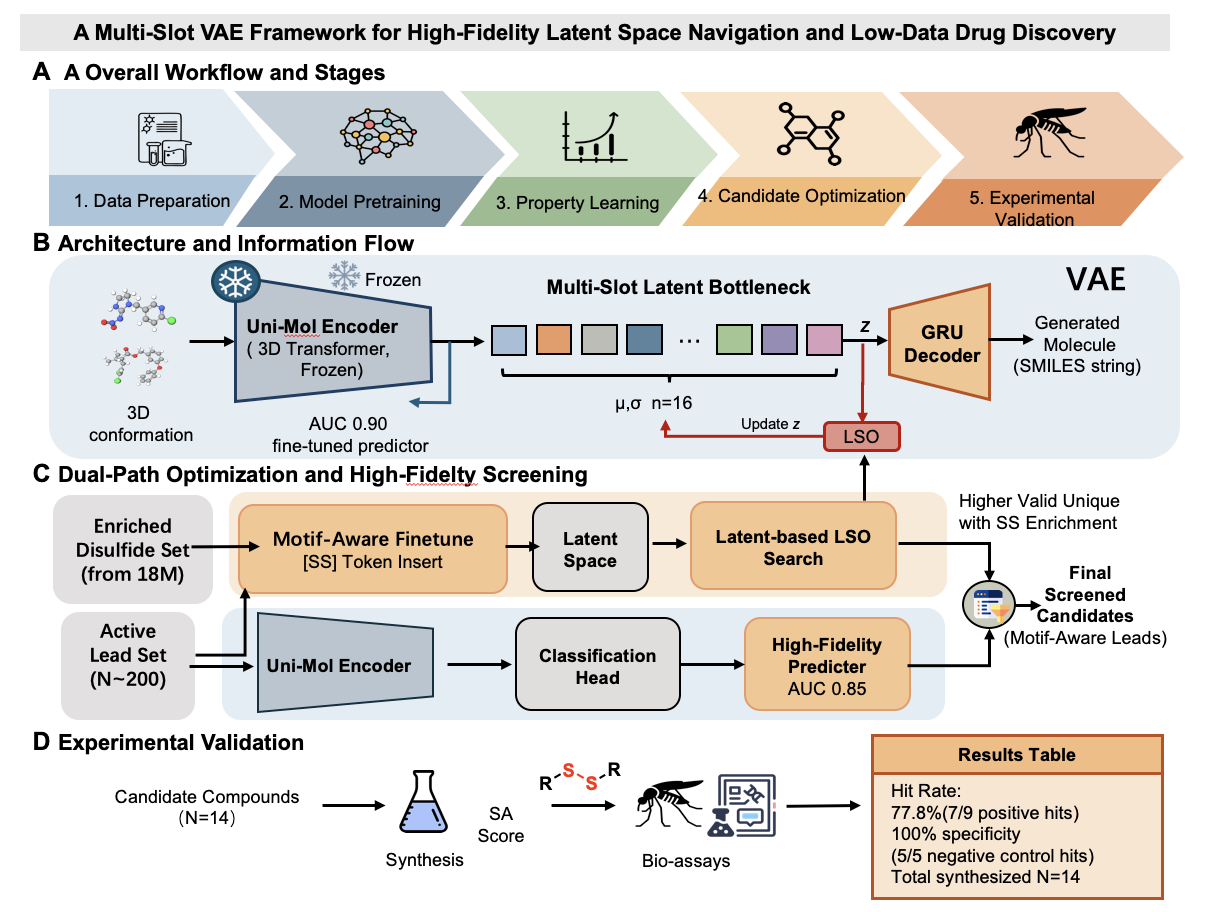} 
\caption{\label{figure_1}\textbf{Schematic overview of the Mos-Gen framework.} A.A Overall Workflow and Stages; B. Architecture and Information Flow; C. Dual-Path Optimization and High-Fidelty Screening ; D. Experimental Validation.}
\end{figure}

\noindent To address these gaps, we propose Mos-Gen: a motif-aware generative framework \cref{figure_1} built upon a high-resolution VAE coupled with a pretrained Uni-Mol\cite{21} encoder, specifically tailored for the design of disulfide-containing mosquito insecticides. The framework employs a dual-path filtering workflow that integrates multi-slot latent space representation, [SS] token-guided structural prior fine-tuning, and high-precision latent space sampling to achieve efficient candidate generation and screening that balances structural fidelity with chemical novelty under extreme label scarcity. Prospective experimental validation demonstrates a 78\% hit rate among predicted positive compounds with only approximately 200 labeled training samples, highlighting the superiority of coupling pretrained structural representations with generative modeling. The main contributions of this work are three-fold: (1) We introduce the motif-aware generative framework specifically tailored for the de novo discovery track of mosquito insecticides; (2) We successfully address the challenge of precisely generating disulfide pharmacophores under extreme data scarcity through a multi-slot latent space architecture and [SS] token-guided structural prior fine-tuning; (3) We demonstrate high-confidence experimental enrichment, achieving a 78\% prospective hit rate with limited labeled training samples.

\section{Methodology}

\noindent \textbf{2.1 Overall Architecture and Dual-Path Workflow}

\noindent To systematically explore the chemical space of disulfide-containing insecticides, we developed a motif-aware generative framework driven by a high-resolution Variational Autoencoder (VAE). The methodology is structured around a "Dual-Path Selection" workflow designed to balance structural fidelity with chemical novelty. 

\vspace{0.5em}

\noindent The pipeline consists of three core phases: (1) structural encoding and baseline generation utilizing a multi-slot latent architecture; (2) motif-aware fine-tuning via token engineering to enforce disulfide constraints; and (3) a high-resolution latent sampling strategy coupled with a fine-tuned predictor for high-fidelity candidate screening. This integrated approach ensures that the generated candidates preserve the essential S-S pharmacophore while expanding the known active scaffold space.

\vspace{1.0em}

\noindent \textbf{2.2 High-Granularity VAE Pretraining}

\noindent The structural backbone of our generative framework leverages a pre-trained Uni-Mol encoder to extract rich 3D conformational features from molecular inputs. To address the challenge of modeling the precise torsional and electronic constraints inherent to the S-S bridge, standard monolithic latent vectors were found to be insufficient. Consequently, we implemented a multi-slot latent representation. The latent space was partitioned into 16 independent, high-granularity subspaces (K=16). This distributed architecture provides the decoder (a Gated Recurrent Unit, GRU) with sufficient degrees of freedom to disentangle and reconstruct complex structural motifs. The baseline VAE was trained using a standard evidence lower bound (ELBO) objective, establishing a continuous and chemically valid latent manifold for downstream optimization.

\vspace{1.0em}

\noindent \textbf{2.3 Motif-Aware Finetuning via Token Engineering}

\noindent To bias the generative manifold toward the specific geometry of disulfide-containing insecticides, we implemented a token-level structural prior. Instead of relying solely on random SMILES decoding, we introduced an explicit [SS] token into the decoder's vocabulary. To embed this semantic constraint into the latent space, we executed a motif-aware fine-tuning protocol. The model was fine-tuned on a diverse background set of approximately 5000 disulfide-containing molecules extracted from a massive 18M unlabeled database and approximately 200 highly relevant lead compounds. This phase forced the model to master the global syntactic rules of S-S bond formation. This fine-tuning strategy successfully anchored the generative process, ensuring that the explicit [SS] token efficiently mapped the latent vectors to the target biologically active chemical space.

\vspace{1.0em}

\noindent \textbf{2.4 Latent Perturbation and High-Fidelity Screening}

\noindent For novel candidate generation, we navigated the optimized latent space using a controlled stochastic perturbation strategy rather than aggressive gradient ascent. We introduced isotropic Gaussian noise \(z^\prime = z + \mathrm{N}(0, \sigma^{2})\)
to the latent representations of the active seed molecules. Empirical analysis revealed a non-monotonic relationship between noise intensity and discovery efficiency. We identified an optimal discovery window at a moderate perturbation scale $\sigma \in [0.05, 0.1]$ . In this regime, the injected variance acts as a thermal activation, facilitating effective scaffold hopping while strictly preventing the collapse of chemical grammar observed at higher noise levels \(\sigma > 0.2\).

\vspace{0.5em}

\noindent Finally, the structurally diverse candidate pool generated within this optimal window was subjected to rigorous filtration. We utilized a fully fine-tuned, representation-based Uni-Mol classification head (Predictor AUC = 0.85) to rank the molecules. This secondary screening ensures that the final motif-aware leads possess both high structural integrity (validity) and a high probability of experimental success.

\section{Results}

\noindent To enable efficient molecular discovery under extremely limited labeled data, we developed a generative–predictive framework built upon a pretrained molecular representation model, Uni-Mol. The workflow integrates large-scale unsupervised representation learning, latent-space molecular generation, and predictor-guided prioritization, followed by experimental validation.

\vspace{1.0em}

\noindent \textbf{3.1 Data-Driven Activity Predictors and Generative Models}

\begin{figure}[h!]
\centering
\includegraphics[width=1\columnwidth]{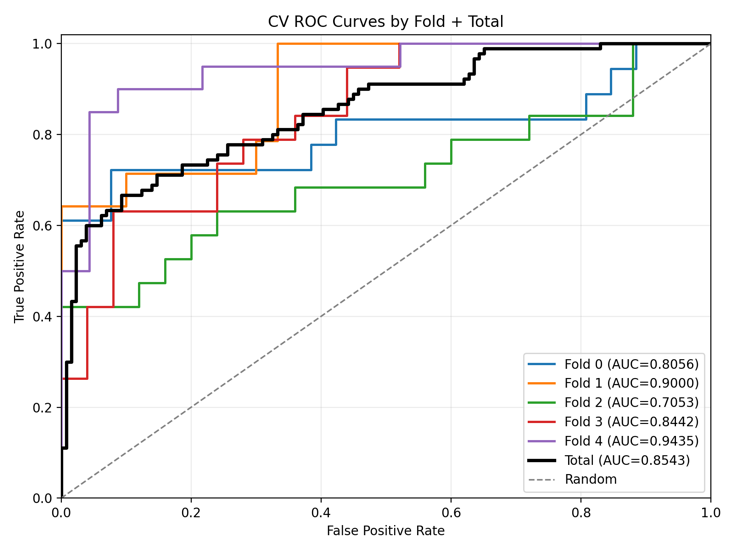} 
\caption{\label{figure_2}. \textbf{Cross-validation performance of the Uni-Mol activity predictor.} Mean AUC = 0.85 across five folds, demonstrating robust generalisation despite a training set of only \(\sim 200\) labled compounds.} 
\end{figure}

\noindent We first established a molecular activity predictor using a dataset of \(\sim 200\) labeled compounds. Despite the limited sample size, the model achieved strong discriminative performance, with a mean AUC of 0.85 under five-fold cross-validation as shown in \cref{figure_2}. This result suggests that the pretrained representation from Uni-Mol provides a robust structural embedding that generalizes well even in low-data regimes. The predictor serves as a key component in downstream candidate prioritization, enabling high-precision filtering from a large pool of generated molecules.

\vspace{1.0em}

\noindent \textbf{3.2 Latent Generative Modeling with Multi-Slot Representation}

\noindent To explore the chemical space beyond the labeled dataset, we constructed a variational autoencoder (VAE) on top of the pretrained encoder. The encoder was kept frozen during training to preserve the structural consistency learned from ~18 million unlabeled molecules.

\vspace{0.5em}

\noindent During the optimization of the generative framework, we discovered that conventional monolithic latent vectors struggled to maintain the precise torsional and electronic constraints inherent to the S–S bridge. To address this, we transitioned to a multi-slot latent representation, which partitions the latent space into multiple independent, high-granularity subspaces.

\vspace{0.5em}

\noindent Our empirical results demonstrated that this distributed latent architecture significantly enhanced the model's capacity to disentangle and reconstruct complex structural motifs. Specifically, increasing the representational granularity allowed the decoder to achieve superior structural fidelity and chemical plausibility, which was essential for capturing the unique pharmacophoric features of the target insecticides. This refined latent formulation provided the necessary resolution to navigate the chemical space effectively, leading to the discovery of structurally diverse candidates that preserve the critical disulfide functionality.

\vspace{1.0em}

\noindent \textbf{3.3 Disulfide-Constrained Structure Generation}

\begin{figure}[h!]
\centering
\includegraphics[width=1\columnwidth]{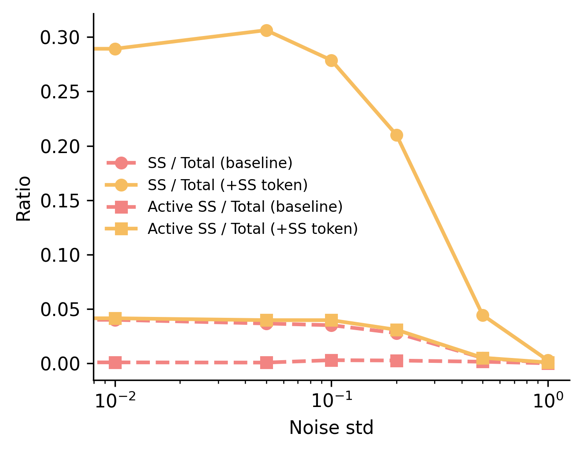} 
\caption{\label{figure_3} \textbf{Comparison of the disulfide-containing and the active disulfide-containing candidate discovery rates between the baseline (without [SS] token) and the motif-aware model (with [SS] token) across a range of latent perturbation scales $\sigma$. }} 
\end{figure}

\noindent To effectively navigate the chemical space of disulfide-containing insecticides, we integrated a domain-specific [SS] token into the VAE vocabulary to serve as a structural prior during decoding. We evaluated the impact of this motif-aware strategy against a standard SMILES-based baseline across a range of latent perturbation scales ($\sigma$).

\vspace{0.5em}

\noindent As illustrated in \cref{figure_3}, the inclusion of the [SS] token significantly enhances the enrichment of target motifs. While the baseline model struggled to generate active disulfide-containing candidates—with the Active SS / Total ratio remaining near zero across all noise levels—our motif-aware approach achieved a substantial and stable discovery yield. This disparity underscores that standard decoder often fails to ground the precise semantic and geometric constraints required for the disulfide pharmacophore. By contrast, the explicit [SS] token facilitates a more efficient mapping between the latent space and the target chemical manifold.

\vspace{0.5em}

\noindent Furthermore, we observed that the generative performance follows a non-monotonic trend relative to the noise intensity. The Active SS discovery efficiency remains optimal within a "Golden Window" of $\sigma \in [0.01, 0.2]$ . In this regime, the model effectively balances structural fidelity with the diversity required for scaffold hopping. When noise exceeds 0.2, both the total SS enrichment and activity yield decline rapidly as the latent vectors drift into adversarial regions, leading to the collapse of chemical grammar. Consequently, we adopted a Dual-Path Selection framework, utilizing moderate latent perturbations $(\sigma \approx 0.1)$ to generate a high-quality candidate pool for downstream screening with our fine-tuned Uni-Mol predictor (AUC 0.85).

\begin{figure}[h!]
\centering
\includegraphics[width=1\columnwidth]{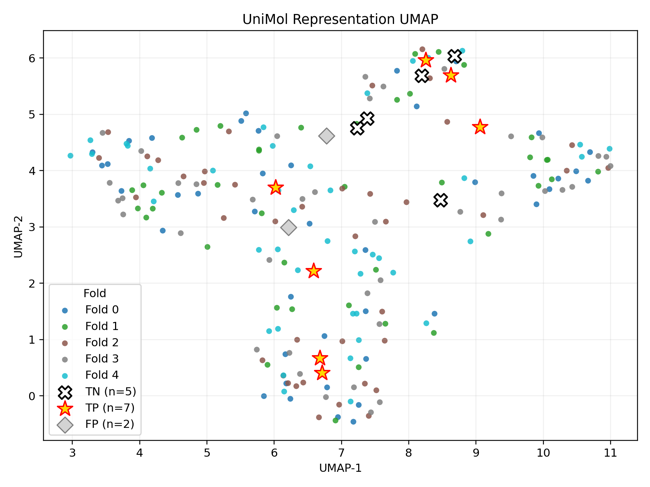} 
\caption{\label{figure_4} \textbf{UMAP visualization of molecular embeddings from the pretrained Uni-Mol model.} UMAP visualization of molecular embedding across 5-fold cross validation with different colors. Active compounds occupy neighboring regions of the embedding space without being exact replicas of training samples, indicating meaningful scaffold variation around known active backbones.} 
\end{figure}

\noindent To examine the relationship between generated candidates and the training data, we visualized molecular representations using UMAP based on the latent embeddings from the pretrained encoder. As shown in \cref{figure_4}, the generated molecules largely reside within the same manifold as the training data.

\vspace{0.5em}

\noindent This observation suggests that the generative model primarily explores locally within the learned chemical space, rather than extrapolating to distant and potentially unreliable regions. Notably, the experimentally validated active compounds are not exact replicas of the training samples, but rather located in neighboring regions of the embedding space, indicating that the model can generate meaningful variants around known active backbones.

\vspace{0.5em}

\noindent While this may limit structural novelty, with limited data, maintaining proximity to the training distribution is preferable to aggressive extrapolation, which often results in chemically invalid or biologically irrelevant candidate compounds.

\vspace{0.5em}

\noindent From an initial pool of generated molecules, a subset of candidates was selected through this multi-stage filtering process, ultimately yielding 14 compounds (9 predicted positives and 5 predicted negatives) for synthesis and experimental evaluation. In addition, basic chemical sanity checks were applied during manual curation to exclude molecules with obviously unstable or hazardous substructures. This generate-then-filter paradigm effectively combines the high recall of generative models with the precision of discriminative models, allowing efficient prioritization under limited experimental budgets.

\vspace{1.0em}

\noindent \textbf{3.4 Experimental Validation}

\noindent Among the predicted positives, 7 compounds exhibited measurable activity, corresponding to a hit rate of 78\%. In contrast, none of the predicted negative compounds showed activity. 

\vspace{0.5em} 

\noindent Considering economic factors, all 14 selected candidate compounds were synthesized via a one-step method. Experimental results highlight the proficiency of the AI-driven model in exploring uncharted chemical space, as 8 out of the 14 selected structures represent entirely novel compounds with no prior literature record.

\vspace{0.5em}

\noindent Overall, these results illustrate that integrating pretrained structural representations, multi-slot latent generative modeling, and conservative latent optimization enables effective molecular design under severe data constraints. Rather than relying on large labeled datasets or highly complex generative strategies, the proposed framework emphasizes stability, interpretability, and practical utility, making it well-suited for real-world drug discovery scenarios where labeled data are often scarce.

\section{Conclusion}

\noindent We presented Mos-Gen, a motif-aware generative framework for the de novo design of mosquito insecticides. By integrating a pretrained Uni-Mol three-dimensional molecular representation, a multi-slot distributed latent space, [SS] token-based structural prior fine-tuning, and a controlled Gaussian latent space perturbation strategy, Mos-Gen enables efficient generation and high-precision screening of active disulfide-containing insecticide candidates. Under an extreme low-data regime of approximately 200 labeled samples, we obtain a 78\% experimental hit rate among predicted positives with no false negatives among predicted negative compounds.

\vspace{0.5em}

\noindent From a broader perspective, this work provides proof-of-concept for introducing generative AI methods into the data-impoverished domain of pesticide and insecticide research. In sharp contrast to the hundreds of thousands of labeled compounds typical in pharmaceutical pipelines, labeled activity data for mosquito insecticides and broader agrochemicals remain orders of magnitude smaller — often numbering only in the hundreds — a data gap that has long constrained the applicability of AI methods. Mos-Gen demonstrates, by fully leveraging pretrained knowledge from large unlabeled chemical databases, domain-specific structural constraint priors, and conservative latent space exploration, that high-precision molecular generation and screening is achievable at such limited label scales, offering a generalizable computational paradigm for the field.

\section{Limitations and Future Work}

\noindent Despite the encouraging experimental results obtained with Mos-Gen, the present study has several limitations that warrant explicit acknowledgment.

\vspace{0.5em}
\noindent First, while the prospective validation of 14 compounds yielded encouraging results, expanding the evaluation to a larger chemical library will further strengthen the statistical robustness and better define the framework's generalized precision. The current 78\% hit rate is computed over 9 predicted positive samples; larger-scale synthesis and testing will be required for a more robust assessment of the framework's true screening precision.

\vspace{0.5em}
\noindent Second, constrained by the sparse training data, UMAP projections indicate that the generated chemical space remains heavily anchored to the local manifold of known actives, thereby limiting far-from-source scaffold hopping. When the objective is the discovery of entirely new scaffold classes rather than proximal variants of known active scaffolds, the coverage capacity of the framework may be insufficient.

\vspace{0.5em}
\noindent Several directions are particularly worthy of pursuit in future work. First, while primary bioassays successfully confirmed mosquitocidal activity in this study, more extensive toxicological evaluations will be systematically conducted to assess practical field applicability. Second, introducing multi-objective optimization modules to simultaneously constrain activity, toxicity, biodegradability, and synthetic accessibility during generation, enabling candidate design more aligned with practical application requirements. Third, transferring the framework to other pesticide discovery tracks (e.g., herbicides, acaricides) to evaluate the generalizability of motif-aware generative methods to low-data agrochemical research more broadly.


\bibliography{example_refs} 
\bibliographystyle{rsc} 










\end{document}